\documentclass[letterpaper]{article} % DO NOT CHANGE THIS
\usepackage{aaai23}  % DO NOT CHANGE THIS
\usepackage{times}  % DO NOT CHANGE THIS
\usepackage{helvet}  % DO NOT CHANGE THIS
\usepackage{courier}  % DO NOT CHANGE THIS
\usepackage[hyphens]{url}  % DO NOT CHANGE THIS
\usepackage{graphicx} % DO NOT CHANGE THIS
\usepackage{natbib}  % DO NOT CHANGE THIS AND DO NOT ADD ANY OPTIONS TO IT
\usepackage{caption} % DO NOT CHANGE THIS AND DO NOT ADD ANY OPTIONS TO IT
\usepackage[titlenumbered,ruled, vlined, linesnumbered]{algorithm2e}
\DontPrintSemicolon

\usepackage{amsmath}
\usepackage{amsfonts}
\usepackage{amssymb}
\usepackage{comment}

\title{Adaptation and Communication in Human-Robot Teaming to Handle Discrepancies in Agents' Beliefs about Plans}
\author {
    Yuening Zhang,
    Brian Williams
}
\affiliations {
    Massachusetts Institute of Technology\\
    zhangyn@mit.edu, williams@mit.edu
}

\begin{document}

\maketitle

\begin{abstract}
When agents collaborate on a task, it is important that they have some shared mental model of the task routines -- the set of feasible plans towards achieving the goals. However, in reality, situations often arise that such a shared mental model cannot be guaranteed, such as in ad-hoc teams where agents may follow different conventions or when contingent constraints arise that only some agents are aware of. Previous work on human-robot teaming has assumed that the team has a set of shared routines, which breaks down in these situations. 
In this work, we leverage epistemic logic to enable agents to understand the discrepancy in each other's beliefs about feasible plans and dynamically plan their actions to adapt or communicate to resolve the discrepancy. 
We propose a formalism that extends conditional doxastic logic to describe knowledge bases in order to explicitly represent agents' nested beliefs on the feasible plans and state of execution.
We provide an online execution algorithm based on Monte Carlo Tree Search for the agent to plan its action, 
including communication actions to explain the feasibility of plans, announce intent, and ask questions. Finally, we evaluate the success rate and scalability of the algorithm and show that our agent is better equipped to work in teams without the guarantee of a shared mental model.

\end{abstract}

\section{Introduction}

When agents collaborate on a task, it is important that they have some shared mental model of the task routines -- the set of feasible plans towards achieving the goals. However, in reality, situations often arise that such a shared mental model cannot be guaranteed. 
For example, in online multi-player games or search-and-rescue missions, people trained separately could form an ad-hoc team where they may follow different conventions. Even if the team has a set of shared routines, novel situations may still occur in which some contingent constraint that forbids certain plans to be taken becomes known only by some agents. In these situations, experienced teammates keep in mind what others know and what actions they may take, and communicate when necessary to make sure the team converges on a feasible plan of action.

Previous work on human-robot teaming, \textit{Pike} \cite{levine2018watching}, assumed that agents share common knowledge of the feasible plans for the task encoded in a knowledge base. Under an equal partner setting, each agent observes the actions taken by others and adapts their actions accordingly, only taking what is still feasible. This approach allows fluid human-robot interaction but breaks down when the common knowledge assumption no longer holds. 

In this work, we generalize the approach to handle situations where there may be discrepancies in agents' beliefs about plans by incorporating  epistemic logic \cite{van2015handbook}, as it provides an explicit representation of agents' nested beliefs towards each other and a mechanism to model communication between agents. 

The contribution of this paper is threefold: (1) We propose the formalism of conditional doxastic logic \cite{baltag2008qualitative} extended to knowledge bases in order to represent agents' nested beliefs on the set of feasible plans for the task and the state of execution. (2) We model both execution and a rich set of communication actions within the framework, including explanation, intent announcement, and question-asking actions, that allows agents to explicitly talk about the feasibility of plans and exchange their intent. (3) We provide an online execution algorithm based on Monte Carlo Tree Search (MCTS) for the agent to dynamically plan its action to adapt to others or communicate to resolve the discrepancy. Finally, we evaluate the success rate and performance of the algorithm through experiments.

\section{Motivating Example}

Consider a pedagogical example where a robot (our agent) and a human collaborate to prepare a drink. The robot has a manipulator arm that can fetch a mug or a glass as the container, and the human can brew some coffee or take some orange juice from the fridge for the drink. For the task to succeed, it must satisfy that: (C1) the mug has to go with the coffee and the glass has to go with the orange juice. 
Under an equal partner setting, from the robot's perspective:
\paragraph{Case1}
If the human doesn't believe constraint C1 holds and thinks that any container can go with any drink, then the robot can adapt to the human by waiting for the human to take the drink first, then fetch the corresponding container. The robot can also explain to the human about constraint C1, especially if the task requires the robot to fetch the container first. The robot can also announce the intent for the human to choose coffee, in which case it can just fetch the mug. 
\paragraph{Case2}
If the human has determined a choice of coffee or juice, but the robot doesn't know which one, the robot may wait for the human to pick first so that it can distinguish their intent, or it can ask the human about their intent.
\paragraph{Case3}
If the human picked up the juice but doesn't know that the robot couldn't reach the glass, the robot may explain the constraint and that the task has failed.

\section{Background}
In order to represent the agents' nested beliefs, our representation builds on top of {conditional doxastic logic} \cite{baltag2008qualitative}, which is one variant in the broader epistemic logic literature. Compared to epistemic logic, it allows the modeling of false beliefs and belief revision by pre-encoding the conditional belief of the agents within the model. Given a set of agents $Ag$ and a set of atomic propositions $At$, 
\textit{conditional doxastic logic} $\mathcal{L}(At, Ag)$  is defined by the following Backus-Naur Form (BNF):
$$\varphi := p |\neg \varphi | (\varphi \land \varphi) |B^{\varphi}_a \varphi,$$
where $p\in At$, $a \in Ag$.
$B^{\psi}_a \varphi$ reads as ``agent $a$ believes $\varphi$ given $\psi$''. Denoting $\top$ as tautology, $B_a\varphi := B^{\top}_a\varphi$ means that agent $a$ believes $\varphi$.
Its semantic model is a \textit{plausibility model}, which is a tuple $M=\langle W, \{\leq_a\}_{a\in Ag}, L\rangle$, where 
\begin{itemize}
    \item $W$: a non-empty set of possible worlds,
    % \item $\geq_a \subseteq W \times W$: a binary relation on $W$ that imposes a plausibility order for agent $a$ 
    \item $\leq_a \subseteq W \times W$: binary relation on $W$ imposing a relative plausibility order between any two worlds for agent $a$,
    \item $L: W \rightarrow 2^{At}$: valuation function mapping each world to the set of atomic propositions that hold in the world.
\end{itemize}
$ w \leq_a v$ means that agent $a$ considers $w$ to be at least as plausible as $v$. $<_a := \leq_a \cap \not \geq_a$ denotes a strict plausibility order. $\simeq_a := \leq_a \cap \geq_a$ denotes an equi-plausibility order. $\sim_a := \leq_a \cup \geq_a$ denotes epistemic indistinguishability, and $cc_a(w) := \{v \in W\ |  w \sim_a^* v\}$ is the set of worlds that agent $a$ finds (possibily more or less) plausible given world $w$, where $\sim_a^*$ is the transitive closure of $\sim_a$. Note that the plausibility relation is reflexive, transitive, locally connected, that is, $v \in cc_a(w)$ implies $v \leq_a w$ or $w \leq_a v$, and well-founded, that is, $min_a(S) := \{w \in S\ | \forall v \in S: v \nless_a w\}$ is well-defined, which is the subset of worlds in $S$ that agent $a$ finds most plausible. A pair $(M, w)$ is a \textit{pointed plausibility model}, which describes a conditional doxastic state with a pointed view at world $w \in W$, i.e. taking $w$ as the true world. The truth of a formula $\varphi \in \mathcal{L}(At, Ag)$ on $(M, w)$, i.e. $(M, w) \vDash \varphi$, can be defined inductively as follows:
\begin{itemize}
    \item $(M, w) \vDash p$ iff $ p \in L(w)$
    \item $(M, w) \vDash \neg \varphi$ iff $M, w \nvDash \varphi$
    \item $(M, w) \vDash \varphi \land \psi$ iff $(M, w) \vDash \varphi$ and $(M, w) \vDash \psi$
    \item $(M, w) \vDash B^\psi_a \varphi$ iff $min_a([\psi]_M \cap cc_a(w)) \subseteq [\varphi]_M$, where $[\varphi]_M := \{w \in W\ | M, w \vDash \varphi \}$ is the set of worlds in $M$ in which $\varphi$ holds.
\end{itemize}

\begin{figure}[!thb]
    \centering
    \includegraphics[scale=0.43]{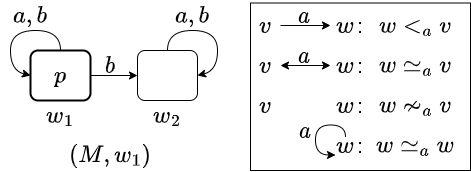}
    \caption{Example pointed plausibility model with legend}
    \label{fig:cdl}
\end{figure}

Figure \ref{fig:cdl} shows an example state represented by a pointed plausibility model $(M, w_1)$ with agents $a$ and $ b$. The two worlds $w_1$ and $w_2$ are labeled with the atomic propositions that hold in the respective worlds. The pointed world $w_1$ highlighted in bold represents the true world in which $p$ holds. The single arrow pointing from $w_1$ to $w_2$ labeled with $b$ indicates that agent $b$ considers $w_2$ to be strictly more plausible than $w_1$. We say $(M, w_1) \vDash B_b \neg p$ since $min_b(cc_b(w_1)) = \{w_2\} \subseteq [\neg p]_M$. If it is instead a double-headed arrow, then it means that agent $b$ considers $w_1$ and $w_2$ to be equally plausible. The lack of any arrow between $w_1$ and $w_2$ for agent $a$ indicates that $w_1 \not\sim_a w_2$, that is, when in $w_1$ or $w_2$, agent $a$ does not consider the other world plausible at all. Since the plausibility relation is reflexive, the self-loops indicate that whichever world it is, the agents find the world plausible. $(M, w_1) \vDash B_a p$ since $min_a(cc_a(w_1)) = \{w_1\} \subseteq [p]_M$, and $(M, w_1) \vDash B_a B_b \neg p$ since $min_a(cc_a(w_1)) = \{w_1\} \subseteq [B_b \neg p ]_M = \{w_1, w_2\}$.

An action is defined by a \textit{plausibility action model} $A = \langle \Sigma, \{\leq_a\}_{a\in Ag} , {pre}, {post}\rangle$, which has a similar structure except instead of a set of worlds $W$, it has a set of events $\Sigma$ representing possible events that may occur in the action. $pre$ and $post$ are functions that assign to each event $\sigma \in \Sigma$ a precondition and a postcondition in $\mathcal{L}(At, Ag)$ respectively, where the postcondition of an event is restricted to a conjunction of literals over $At$ or $\top$. A \textit{pointed plausibility action model} is a pair $(A, \sigma)$, $\sigma \in \Sigma$, which describes an action where $\sigma$ is the true event.

In general, a pointed plausibility model for state or action can point at multiple worlds, such as $(M, W_d)$ or $(A, \Sigma_d)$. $W_d$ and $\Sigma_d$ are called the \textit{designated} worlds or events. For example, given a state $(M, w)$, $(M, W_d)$ with $W_d = min_a(cc_a(w))$ represents agent $a$'s local perspective of the state, where $W_d$ includes all the worlds that agent $a$ finds the most plausible. $(M, W_d)$ is a \textit{global} state if $|W_d| = 1$. Additionally, $(M, W_d) \vDash \varphi$ iff $(M, w) \vDash \varphi$ for all $w \in W_d$. An action $act$ updates a state $s$ through \textit{action-priority update} $s \otimes act$, which we refer the readers to the details in \cite{baltag2008qualitative, bolander2011epistemic}.

\section{Approach Overview}
Our solution requires answering three questions: (1) what representation to use to capture the agents' nested beliefs of the set of feasible plans and the state of execution, (2) how to model execution and communication actions and how they update the state, and (3) how to strategically choose the next action.
Our key insight is to extend conditional doxastic logic to describe knowledge bases, and use the knowledge bases to encode the feasible plans and state of execution, so that we can describe agents' beliefs on the plan space instead of their beliefs on state. As a result of this new logic, execution and communication actions can be defined which operate by adding or removing constraints from the knowledge bases. With the state and action models defined, we use an MCTS-based algorithm to simulate forward in the next $k$-step horizon to decide what is the best action to take.

\begin{figure}[!thb]
    \centering
    \includegraphics[scale=0.23]{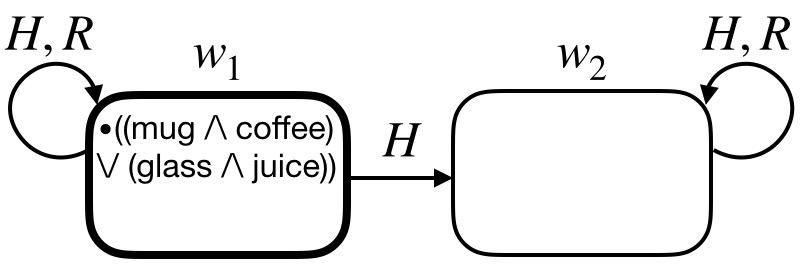}
    \caption{Plausibility model for nested beliefs on plans}
    \label{fig:approach_1}
\end{figure}

\begin{figure}[!thb]
    \centering
    \includegraphics[scale=0.23]{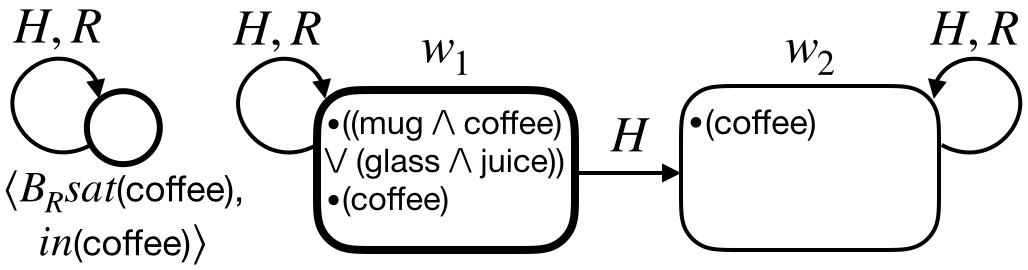}
    \caption{intent announcement (left) \& resulting state (right)}
    \label{fig:approach_4}
\end{figure}

For example, Figure \ref{fig:approach_1} captures the agents' nested beliefs on plans from {Case1}. Each world in the plausibility model is now a knowledge base that contains the constraints of the task. Since $H$ (human) finds $w_2$ more plausible, $H$ believes that constraint C1 does not need to hold. An example action where the robot announces the intent of coffee is shown in Figure \ref{fig:approach_4} (left). The action has a single event whose precondition is that $R$ (robot) must believe that adding the constraint of coffee is satisfiable given its belief of the current feasible plans. As a result of the action, all worlds now have the constraint of coffee added, including $w_2$ that $H$ believes in.

\section{Representing Team's Nested Beliefs on Plans}

We describe our representation in two parts: (1) conditional doxastic logic on knowledge bases and its semantics, (2) our task representation and its encoding in the knowledge base.

\subsection{Conditional Doxastic Logic on knowledge Bases}

Given a finite set of atomic propositions $At$, and a finite set of agents $Ag$, \textit{conditional doxastic logic on knowledge bases} $\mathcal{L}_{KB}(At, Ag)$ is defined by the BNF:
$$\varphi := in(c)| entailed(c) |\neg \varphi | (\varphi \land \varphi) |B^{\varphi}_a \varphi, $$
in which $a \in Ag$, $c \in \mathcal{C}(At)$, where $\mathcal{C}(At)$ is the classical propositional logic $c := p |\neg c | (c \land c)$, $p \in At$. Note that the formulation naturally extends to constraint systems with finite-domain variables, which is what we use. We hence refer to $c$ as a constraint. $in(c)$ means constraint $c$ is an explicit member of the constraints in the knowledge base, $entail(c)$ means constraint $c$ is entailed by the knowledge base, and we define $sat(c) :=  \neg entailed( \neg c)$, which means constraint $c$ is satisfiable by the knowledge base. 

The plausibility model for $\mathcal{L}_{KB}(At, Ag)$ is a tuple $M=\langle W, \{\leq_a\}_{a\in Ag} , KB\rangle$, {where} $W$ and $\leq_a$ are the same as before and $KB: W \rightarrow \mathbf{KB}_{\mathcal{C}(At)}$ is a function that maps each world to an associated knowledge base in $\mathcal{C}(At)$.
When determining the truth of a formula $\varphi \in \mathcal{L}_{KB}(At, Ag)$ on a pointed plausibility model $(M, w)$, we replace the first rule on $(M, w) \vDash p$ in the inductive rules with the following:
\begin{itemize}
    \item $(M, w) \vDash in(c)$ iff $ c \in KB(w)$
    \item $(M, w) \vDash entailed(c)$ iff $KB(w) \vDash c$ 
\end{itemize} 

We can say the following about the state in Figure \ref{fig:approach_1}:

\begin{itemize}
    \item $B_R in((\mathtt{mug} \land \mathtt{coffee})) \lor (\mathtt{glass} \land \mathtt{juice}))$
    \item $B_R B_H \neg in((\mathtt{mug} \land \mathtt{coffee})) \lor (\mathtt{glass} \land \mathtt{juice}))$
    \item $\neg B_R entailed(\mathtt{mug} \land \mathtt{coffee})$
    \item $B_R \neg sat(\mathtt{mug} \land \mathtt{juice}) \land B_R B_H sat(\mathtt{mug} \land \mathtt{juice})$

\end{itemize}

\subsection{Task Representation \& Encoding}

The set of feasible plans towards achieving the goals of the task forms a \textit{plan library} for the task. Additionally, actions may be ordered in the plan, such as requiring the container to be picked up first before the drink. Therefore, our task representation is a \textit{temporal plan library} $\langle V, E, O, C\rangle $, where:
\begin{itemize}
    \item $V$ is a set of decision variables with $domain(v), v\in V$.
    \item $E$ is a set of time points with guard condition $guard(e)$ for each $e \in E$, a conjunction of decision variable assignments. 
    $e$ should be executed iff $guard(e)$ is satisfied.
    \item $O$ is a set of ordering constraints $o = \langle e_i, e_j, guard(o)\rangle$, $o \in O$, requiring time point $e_i$ to precede time point $e_j$ in execution order if its guard condition $guard(o)$
    is satisfied. We assume $guard(o) \vDash guard(e_i) \land guard(e_j)$.
    \item $C$ is a set of constraints scoped on $V$.
\end{itemize}

The time points represent the actual events of taking the actions. In multi-agent case, a \textit{multi-agent temporal plan library} $\langle V, E, O, C, Ag, f\rangle $ additionally has a set of agents $Ag$ and a function $f: E \rightarrow Ag$ that maps each time point to an agent that it belongs to. In our formulation, the decision variables do not have ownership. This reflects our equal partner setting in which decisions do not belong to any agent and an announced intent can affect multiple agents' actions. 

The plan library represents a set of \textit{candidate subplans} $G$, where a subplan $g \in G$ is a full assignment to all the decision variables $V$. We use $E_g$ and $O_g$ to denote the set of time points and ordering constraints activated by $g$, i.e. those whose guard conditions are satisfied. A subplan induces a set of total orderings on $E_g$ that satisfies $O_g$, which we denote by $T_g$. A subplan $g$ is \textit{feasible} iff all the constraints $C$ are satisfied, i.e. $\forall c \in C$, $g \vDash c$, and there exists a total ordering of $E_g$ that satisfies $O_g$, i.e. $T_g \neq \emptyset$.

\paragraph{Execution} As execution progresses, decision variables are gradually grounded either implicitly from the execution of time points or explicitly from announcement of intent. The \textit{execution state} is a tuple $\langle t, C_I \rangle$, where $t$ is an \textit{execution history}, which is a total ordering of time points $(e_i, e_j, ..., e_k)$ that have been executed, and $C_I$ is the set of intents that have been announced during execution. An intent, in its most general form, can be an arbitrary constraint scoped on $V$, but is commonly an assignment to a specific decision variable.  The subplans that are \textit{feasible with respect to} $\langle t, C_I \rangle$ include any feasible subplan $g$ such that there exists $t_g \in T_g$, where $t$ is the prefix of $t_g$, and $g$ satisfies $C_I$. Execution \textit{fails} when there exists no feasible subplan with respect to $\langle t, C_I \rangle$. Execution \textit{succeeds} when there exists a feasible subplan $g$ with respect to $\langle t, C_I \rangle$ such that $t \in T_g$. Note that execution can succeed without ever converging to a unique subplan, and it is possible for further time points to be executed and move away from the success state. 

\paragraph{Encoding in Knowledge Base}
We encode the plan library and the execution state in the knowledge base, so that at any point during execution, the knowledge base contains all the feasible subplans with respect to $\langle t, C_I \rangle$. We ensure that the knowledge base is consistent iff execution has not failed.

The variables of the encoding include (1) a discrete variable for each decision variable $v_i \in V$ with the same domain $domain(v_i)$, and (2) a boolean variable for each time point $e_i \in E$ with domain $\{\mathtt{T}, \mathtt{F}\}$, representing if the time point is executed. We add the following constraints to the knowledge base prior to execution:
\begin{itemize}
    \item The constraints $C$ as defined in the plan library.
    \item For each time point $e_i$, $((e_i = \mathtt{T}) \rightarrow guard(e_i))$, i.e. if time point $e_i$ is executed, its guard condition must hold.
    \item Negation of \textit{nogoods} \cite{katsirelos2005generalized} that represent any combination of choices of $V$ that would result in an inconsistent ordering of time points. This can be computed from the ordering constraints $O$.
\end{itemize}
During execution, we may additionally add to the KB:
\begin{itemize}
    \item Announced intents $C_I$.
    \item For each execution of time point $e_j$, a conjunction of (1) assignment of variable $e_j$ to $\mathtt{T}$, and (2) the negation of the guard condition $\neg guard(o)$ for any ordering constraint $o = \langle e_i, e_j, guard(o) \rangle$ in which the predecessor $e_i$ has not been executed by the time $e_j$ is executed.
\end{itemize}

The last rule ensures that for any ordering constraint $o  = \langle e_i, e_j, guard(o) \rangle$, 
if the guard condition holds, then if $e_j$ is executed, $e_i$ must also have been executed, hence satisfying the ordering constraint.
Note that we only encode the set of time points that have been executed, instead of their actual order of execution. 
With the above encoding, given a knowledge base KB, execution fails iff KB $ \vDash \bot$. Execution succeeds iff there exists subplan $ g$, i.e. a full assignment of $V$, such that KB$ \land g \nvDash \bot$ and $\forall e_i \in E_g$, KB $ \vDash (e_i = \mathtt{T})$. We denote the success condition by $suc_{(V, E)}$, and say that execution succeeds iff KB $ \vDash suc_{(V, E)}$. Additionally, $(M, w) \vDash suc_{(V, E)}$ iff $ KB(w) \vDash suc_{(V, E)}$.

Take Case 1 as an example, the knowledge base contains discrete variables $container$ with domain $\{\mathtt{mug}, \mathtt{glass}\}$ and $drink$ with domain $\{\mathtt{coffee}, \mathtt{juice}\}$, and boolean variables $e_{mug}$, $ e_{glass}$, $ e_{coffee}$, $ e_{juice}$ representing the events of picking up each item. Using $\mathtt{mug}$ as a shorthand for $(container=\mathtt{mug})$ and similarly for others for the purpose of decluttering, the constraints include:
\begin{enumerate}
    \item $(\mathtt{mug} \land \mathtt{coffee}) \lor (\mathtt{glass}\land \mathtt{juice})$
    \item $(e_{mug}=\mathtt{T}) \rightarrow  \mathtt{mug}$, similarly for other time points
\end{enumerate}

Note that we use the same shorthands throughout the rest of the paper. 
In this example, when the robot picks up the mug, constraint $(e_{mug} = \mathtt{T})$ is added to the knowledge base. From 2 above, we now have KB $ \vDash  \mathtt{mug}$, and consequently from 1, we have KB $ \vDash \mathtt{coffee}$, which limits the human's choice of drink to coffee. Picking up juice is no longer feasible since KB $ \land (e_{juice} = \mathtt{T}) \vDash \bot$. Consider another case where the robot's action must precede the human's action, i.e. there are ordering constraints $\langle e_{mug}, e_{coffee}, \mathtt{mug} \land \mathtt{coffee}\rangle$, $\langle e_{glass}, e_{coffee}, \mathtt{glass} \land \mathtt{coffee}\rangle$, etc. If the human picks up the coffee before the robot takes any action, then $(e_{coffee} = \mathtt{T}) \land \neg ( \mathtt{mug} \land \mathtt{coffee}) \land \neg ( \mathtt{glass} \land \mathtt{coffee})$ is added, resulting in an inconsistent knowledge base. 

\begin{comment}
Need to prove Theorem 1, using Lemma on comment 2/3

Basically, KB includes the candidate subplans that are still feasible wrt the partial execution. 

Why? Intuitively, C is satisfied as explained, Ordering is satisfied as explained. partial execution <G', T'> is enforced by the event guard constraints.
    
\end{comment}

\begin{comment}
\textit{Execution hard failure}, which is failure that cannot be fixed, can be determined by checking that $KB \vDash \bot$.
\end{comment}

\section{Dynamic Model of Evolution}

In this section, we describe how the model evolves as a result of execution or communication actions. We first introduce the plausibility action model for our extended logic, then describe how to model each type of action. In this work, we assume that agents observe all actions that are taken, that is, all actions are public. 
\subsection{Plausibility Action Model for Knowledge Bases}

A \textit{plausibility action model} $A$ for $\mathcal{L}_{KB}(At, Ag)$ is a tuple $\langle \Sigma, \{\leq_a\}_{a\in Ag} , {pre}, {post}\rangle $, {where} $\Sigma$ and $\leq_a$ are the same, and $pre$ and $post$ are functions that map each event to a formula in $\mathcal{L}_{KB}(At, Ag)$. The postcondition is restricted to a conjunction of $in(c)$, which adds constraint $c$ to the knowledge base, and $\neg in(c)$, which removes constraint $c$ from the knowledge base if it exists, as well as $\top$, i.e. nothing changes. For this paper, we further restrict the postcondition to be either $in(c)$ or $\top$, i.e. adding at most one constraint to the knowledge base. The \textit{action-priority update} updates the knowledge bases as described accordingly.

\begin{figure}
    \centering
    \includegraphics[scale=0.22]{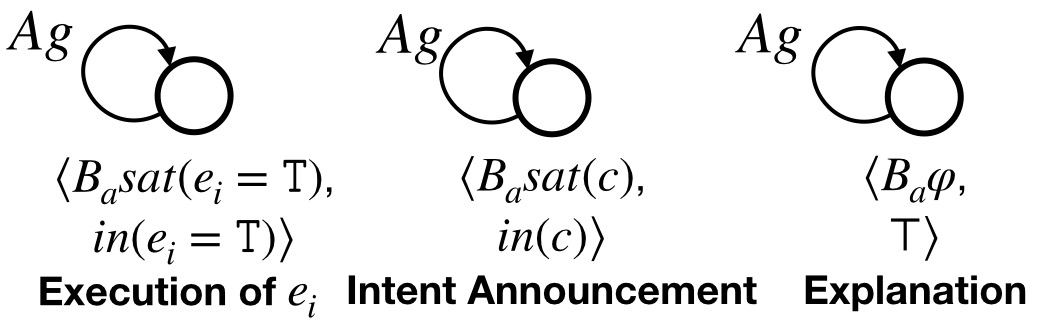}
    \caption{Action representations}
    \label{fig:template}
\end{figure}

\paragraph{Execution Action}
An {execution action} is the action of an agent executing a {time point}, such as robot picking up mug. Recall that in our setting, each time point is assigned to an agent who can execute it. Given that the time point being executed is $e_i \in E$, and the agent who executes it is $a = f(e_i)$, the simplest case of execution of time point $e_i$ that has no potential predecessors is shown in Figure \ref{fig:template} (left).
We assume agents are rational and for agent $a$ to execute time point $e_i$, it needs to believe that executing $e_i$ is feasible, i.e. $B_a sat(e_i = \mathtt{T})$. All agents observing the action also observe the truth of agent $a$ having such belief. For the postcondition, as $e_i$ is executed, the constraint $(e_i =\mathtt{T})$ is added to the knowledge base.

\begin{figure}
    \centering
    \includegraphics[scale=0.186]{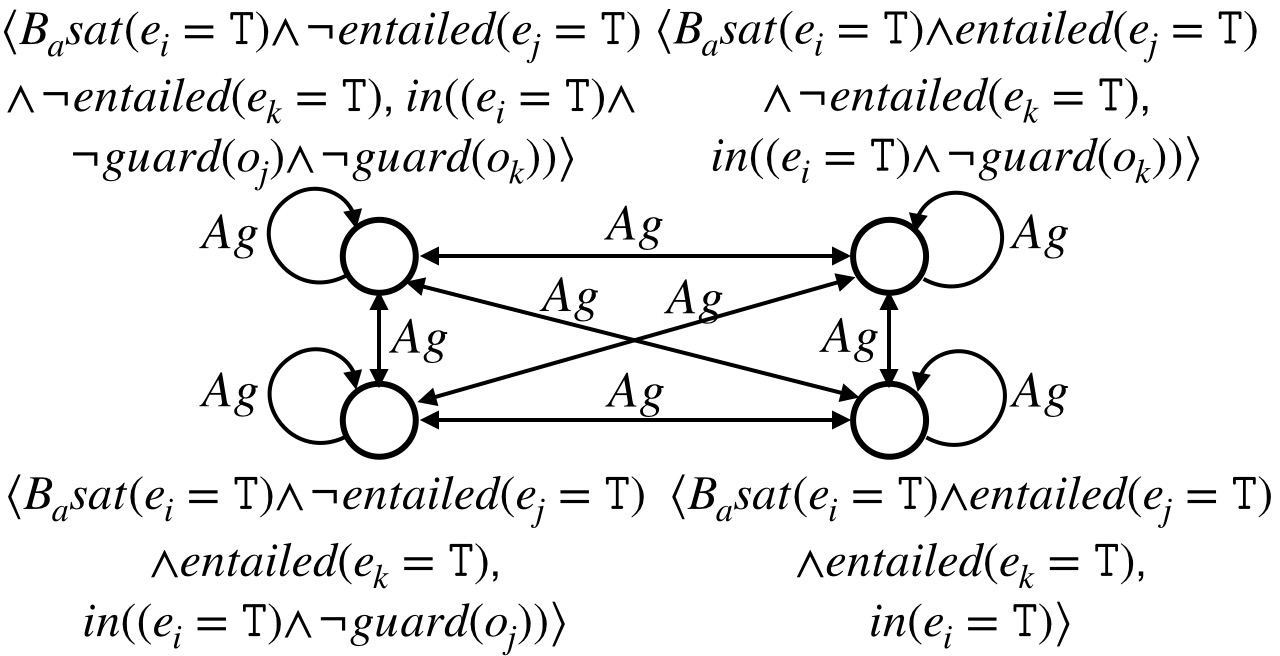}
    \caption{Example execution action of $e_i$ with ordering constraints $\langle e_j, e_i, guard(o_j) \rangle$ and $\langle e_k, e_i, guard(o_k) \rangle$}
    \label{fig:execution_2}
\end{figure}

When there are potential predecessors for $e_i$, we need to make sure the corresponding ordering constraints are satisfied. 
The postcondition of the event should always add $(e_i = \mathtt{T})$ to the knowledge base, and for any ordering constraint $o = \langle e_j, e_i, guard(o) \rangle$, add $(\neg guard(o))$ on condition that $\neg entailed( e_j = \mathtt{T})$, that is, if $e_j$ has not been executed. While a more succinct action model specification is possible \cite{Ditmarsch2006SemanticRF}, we use the standard form defined above by taking the cross product of all the predecessors, and creating equi-plausible events for them as shown in Figure \ref{fig:execution_2}.  Even though the size of the action model is exponential to the number of potential predecessors each time point has, because the preconditions of these events are mutually exclusive, the model size for the updated state will not increase as a result of the action update.

\paragraph{Intent Announcement}
The model for an intent announcement action is shown in Figure \ref{fig:template} (middle). For agent $a$ to announce the intent, it must believe that it is satisfiable, hence the precondition $B_a sat (c)$. The intent is added as a postcondition.
Figure \ref{fig:approach_4} shows an example intent announcement.

\paragraph{Explanation}
The model for an explanation action where agent $a$ explains its belief of $\varphi \in \mathcal{L}_{KB}(At, Ag)$ is shown in Figure \ref{fig:template} (right). The precondition says that agent $a$ has to believe $\varphi$, i.e. agents cannot lie about their belief. To the other agents, the explanation is essentially a public announcement that agent $a$ believes $\varphi$. This means that whether a particular agent adopts the explainer's belief depends on the conditional belief pre-encoded in the initial pointed plausibility model, which specifies how an agent's belief gets revised when a new piece of evidence is received.

\begin{figure}[!thb]
    \centering
    \includegraphics[scale=0.23]{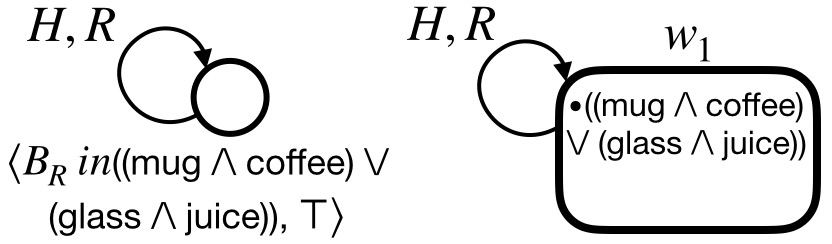}
    \caption{Explanation action (left) and resulting state (right)}
    \label{fig:approach_2}
\end{figure}

An example explanation action where the robot explains constraint C1 is shown in Figure \ref{fig:approach_2}. Based on the pointed plausibility model in Figure \ref{fig:approach_1}, upon the announcement that the robot in fact believes C1 holds, $w_2$ is eliminated as it does not satisfy the precondition, and the human is left with $w_1$ in which C1 holds. Depending on the initial conditional belief, it is also possible to have situations where the human does not trust the robot and does not adopt its belief.

In this paper, we restrict the explained formula to be of the BNF form $\varphi := \neg \varphi |B_a \varphi | in(c) $, where $a \in Ag$, $c \in \mathcal{C}(At)$. This simplifies the explanations in that (1) the explained formula cannot be arbitrarily complex such as $B_a in(c) \rightarrow B_b in(c)$, (2) the explanation must be about whether the knowledge base contains a constraint or not, instead of the satisfiability or entailment of an arbitrary constraint. This is similar in spirit to the idea of abductive explanations, where we want to give an explanation $c$ such that together with the existing theory $T$, it explains an explanandum $O$, i.e. $T \cup \{c\} \vDash O$. In this case, what is satisfiable or entailed is often the explanandum, and what constraints should or should not be in the theory is what we explain.

\paragraph{Question-Asking}

\begin{figure}[!thb]
    \centering
    \includegraphics[scale=0.21]{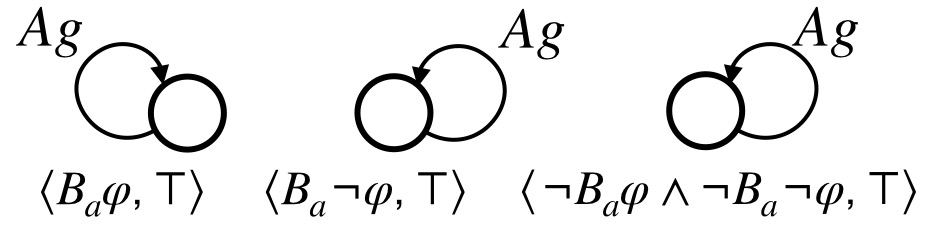}
    \caption{Question-asking action}
    \label{fig:ask_1}
\end{figure}

An agent can ask another agent about something that it is uncertain of. Since we assume public actions, the answer is observed by all agents. Given that agent $a$ is asked about its belief on formula $\varphi \in \mathcal{L}_{KB}(At, Ag)$, the pointed plausibility action model is shown in Figure \ref{fig:ask_1}. We place the same restriction on $\varphi$ as in the explanation actions.

\begin{figure}[!thb]
    \centering
    \includegraphics[scale=0.23]{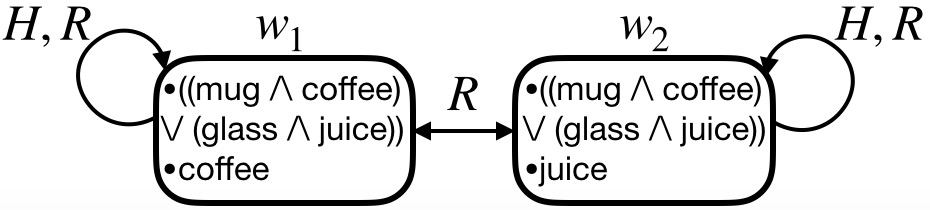}
    \caption{Example pointed plausibility model where the robot is uncertain about the human's choice}
    \label{fig:ask_2}
\end{figure}

Using Case 2 as an example, the robot does not know which choice of drink the human has determined on, which is represented by the pointed plausibility model in Figure \ref{fig:ask_2}. The robot can ask a question about the human's belief on $in(\mathtt{coffee})$, i.e. whether its intent is to take coffee. The resulting state would have the double-headed arrow in the middle labeled with $R$ removed, i.e. the robot will be able to distinguish human's intent.

\section{Online Execution Problem}

We assume that execution is asynchronous, all actions are public, and communication has a cost. We also assume the discrepancies in beliefs come only from the agents' initial beliefs on constraints $C$, i.e. they share the belief on the rest of the plan library such as the ordering constraints and the guard for the time points. In this paper, we assume that a task involves two agents (e.g. robot and human), though there is no theoretical barrier to applying it to more agents.
The online execution problem from a single agent's perspective, say agent $a$, involves taking the input of the following prior to execution:
\begin{itemize}
    \item A multi-agent temporal plan library $\langle V, E, O, \{\}, Ag, f\rangle$.
    \item A pointed plausibility model $s_0 = (M, W_d)$ capturing agents' initial nested beliefs on constraints $C$ from agent $a$'s perspective, such that $W_d = cc_a(w), \forall w \in W_d$.
\end{itemize}
Note that constraints $C$ is empty in the plan library as it is captured by the input of $s_0$. $W_d$ includes any world that agent $a$ finds plausible (not necessarily most plausible), and we assume that $(M, W_d)$ captures the ground truth state $(M, w^*)$ as one of its possibilities, so that the agent's belief can also be revised if needed. During execution, the agent receives the input of a stream of actions that are taken by itself or others in real-time, including execution and communication actions. Each action triggers a callback and the agent outputs an action to be taken or none.

\paragraph{Overall Algorithm}

Upon receiving an action, our agent determines how it should act next -- either take an execution action, communicate, or wait for others to act. It simulates forward to predict the utility of each possible action, e.g. if others may follow up with incorrect actions, or if many communication actions will be needed. The algorithm draws insight from epistemic planning for implicit coordination \cite{engesser2017cooperative} and relies on the agent's ability to take others' perspectives to predict their actions.

The overall algorithm is illustrated in Algorithm \ref{alg:main}, which we name it \textit{Epistemic Pike} (\textit{EPike}) after {\textit{Pike}} \cite{levine2018watching}. 
Prior to execution, the agent compiles the initial state $s$ from $s_0$ and the plan library as described in the knowledge base encoding section (line 1). Upon receiving an action $act'$, the updated state (line 3) is checked for several conditions. If the agent believes that execution has failed (line 4), then it explains the failure when some agent might not know (line 5 - 6). If the agent is unsure about whether execution has failed (line 7), then it sees if it can ask someone to distinguish it (line 8). If both conditions do not apply, then execution has not failed, and the agent checks if execution has succeeded (line 9). If so, then the agent explains it when some agent might not know (line 10 - 11). If execution has not succeeded, then the agent searches for the next action to take, if any, to progress toward completing the task (line 13). Note that when a question-asking action is taken, we wait until the answer is announced before encoding the answer as an explanation action that gets observed.

\begin{algorithm}
	\caption{Online Execution for Ego Agent $a$}
	\label{alg:main}
	\SetKwInOut{Input}{Input}
	\SetKwInOut{Output}{Output}
	\Input{$V$, $E$, $O$, $Ag$, $f$, $s_0 = \langle M, W_d \rangle$, agent $a$, \\ {Online: } $act'$, an observed action }
        
	\Output{{Online: } $act$, an action or $None$}
        \textbf{Offline: }
        $s \leftarrow \textsc{CompileInitialState}(s_0, V, E)$\;

        \textbf{Online upon observing $act'$:}\\
        $s \leftarrow s \otimes act'$\;
        \If{$s \vDash B_a entailed(\bot)$} {
            % \tcp{Execution failed}
            \If{$s \nvDash  B_a (\land_{i\in Ag} B_i entailed(\bot))$}{
                \Return $\textsc{ExplainFailure}(s)$\;
            }
        } 
        \ElseIf { $s \nvDash B_a \neg entailed(\bot)$}{
            % \tcp{Uncertain if execution failed}
            \Return    $\textsc{AskIfFailure}(s)$\;
            }
        \ElseIf{$s \vDash B_a suc_{V, E}$} {
            % \tcp{Execution satisfies termination condition}
            \If{$s \nvDash B_a (\land_{i\in Ag} B_i suc_{V, E})$}{
                \Return $\textsc{ExplainSuccess}(s)$\;
            }
        }
        \Else{\Return $\textsc{SearchAction}(s)$\;}
        \Return $None$;
      
\end{algorithm}

Each online subroutine in Algorithm \ref{alg:main} calls an MCTS algorithm with a different configuration. The MCTS algorithm simulates the team's possible execution in the next $k$-step horizon, and based on the result of the simulations, the agent decides if it should take an action now and which action to take. Our MCTS algorithm can be configured on (1) the termination conditions, including the horizon $k$, (2) which types of actions to consider for both the ego agent and the other agents, and (3) the penalties for communication actions, since we assume communication has a cost. We add a fifth type of action, \textit{noop}, to represent agent taking no action and waiting for others to act. 

For \textsc{SearchAction}, a node terminates if its state $s$ satisfies either $s\vDash entailed(\bot)$, which gives a utility of 0 (execution fails), or $s \vDash suc_{(V, E)}$, which gives a utility of 1 (execution succeeds), or if simulation reaches a horizon of $k$, which gives a utility of 1 (execution has not failed). Note that only execution actions increment the horizon, 
since we care about the outcome after the next three physical actions. During search, we consider all five types of actions (including noop) from all agents, except for the intent announcement and question-asking actions from the other agents. They can be reasonably omitted to reduce the search space, since they may be unpredictable and ignoring it does not prevent the simulated execution to reach success state. 

For the rest of the subroutines, a node terminates if its state $s$ satisfies $s\vDash \land_{i \in Ag} B_{i} entailed(\bot)$ for \textsc{ExplainFailure}, $s\vDash B_{a} entailed(\bot)\lor B_{a} \neg entailed(\bot) $ for \textsc{AskIfFailure}, and $s\vDash \land_{i \in Ag} B_{i} suc_{(V, E)} $ for \textsc{ExplainSuccess}, all giving a utility of 1. For \textsc{ExplainFailure} and \textsc{ExplainSuccess}, only explanation and question-asking actions of the ego agent are considered. Asking a question may still be useful if the agent is uncertain about what others currently believe. For \textsc{AskIfFailure}, only question-asking actions for the ego agent are considered. In these cases, since the ego agent is just looking to inform others or ask a question, it is reasonable to ignore what other agents may do. To penalize communication, we set a penalty factor of 0.9 for explanation actions and question-asking actions, and 0.85 for intent announcement actions, though the values may change depending on applications.
Penalty is a multiplication factor to the utility of the node. Execution actions and noop action are not penalized.

\begin{figure}
    \centering
    \includegraphics[scale=0.37]{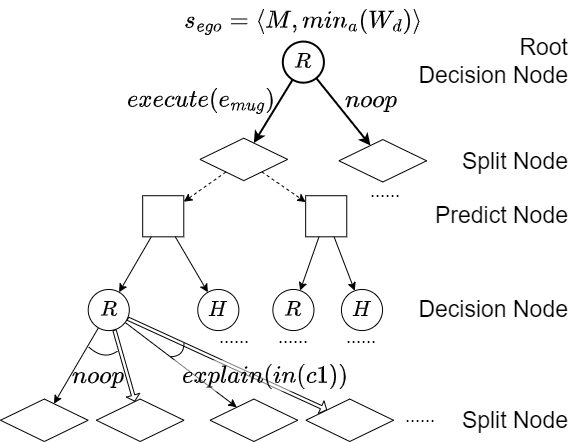}
    \caption{Partially expanded search tree}
    \label{fig:tree}
\end{figure}

\paragraph{Search Tree} We describe the expansion of the search tree using $\textsc{SearchAction}$ as an example, before discussing the details of MCTS. A partially expanded search tree is shown in Figure \ref{fig:tree}. There are four types of nodes in the tree: root decision node (bold circle), split nodes (diamonds), predict nodes (squares), and decision nodes (circles). Each node has its state $s_i$ and has a utility score of between $[0, 1]$. 

The root decision node is only used as the root of the tree and finds the best action to take for the ego agent (including noop). Given input of $s = (M, W_d)$ from  the subroutine, the state of the root decision node is the ego agent's current belief of the state $s_{ego}=(M, min_a(W_d))$. The node branches on all the possible actions the ego agent can take based on its current belief, creating children split nodes. We discuss the generation of possible actions in the Appendix. If there exist children with positive scores, the agent chooses the action that leads to the child with the maximum score, and prefers non-noop actions when there is a tie.

The split node represents the state after the application of the action, which may point at multiple worlds. The split node splits the state into a set of global states where only one unique world is pointed at, and answers the question of: of all the possible states that the action can lead to, what is the worst-case situation that can happen. 

Each predict node predicts what may happen from the global state. To do so, it expands into a set of decision nodes for the agents and predicts how each agent contributes to progressing the state toward success. If the parent split node results from an agent taking noop, then the predict node does not expand on the decision node for the same agent, that is, the agent has to wait for someone else to take action.

Each agent's decision node expands on the set of possible actions the agent can take. Assuming the parent predict node has state $s= (M, w)$, this will be the set of actions that the agent finds feasible from its perspective  $(M, min_a(cc_a(w)))$. For each action, we expand on it both from the agent's subjective view of the state (a thick arrow) to determine how good the action is from the agent's perspective, and from the objective view of the state (a single arrow), i.e. the same perspective as the parent predict node, to determine how good the action actually is. The root decision node can be considered as a special decision node where the subjective and objective views are the same. We assume that the agent only takes the best actions from its perspective, i.e. the ones with the highest subjective score that is greater than 0, and has a uniform probability of choosing any action from that set, with the exception that if there exist perfect execution actions with subjective scores of 1, then the agent would not consider taking noop action.
For each node, we can determine what perspective the state is viewed from by traversing the thick arrows from the root, which represent perspective shifts. A node reaches termination state if it satisfies the termination condition defined earlier.

\begin{figure*}
  \centering
    \includegraphics[scale=0.49]{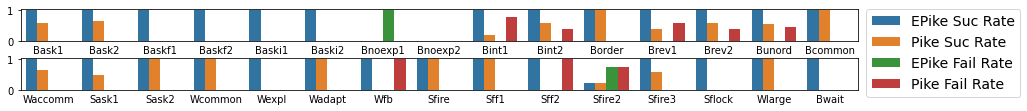}
    \caption{Success rate and failure rate of EPike and Pike on a set of hand-crafted test cases}
    \label{fig:exp_3}
\end{figure*}

\paragraph{Tree Policy \& Simulation (Default) Policy}
Regarding the tree policy, for each node, we compute the UCB1 score of the children to select which one to descend down the tree. For the split node, we use the negative score of each child as the exploitation term, to prioritize simulations of the worst-case situation. For the decision node, we use the subjective score of each action as the exploitation term, to prioritize simulations of the actions that are likely taken by the agent. Once the action is selected, out of the two children split nodes from the objective view and the subjective view, we select the one that is less expanded. 

For the simulation policy, at each decision node, we only consider the execution actions of the agent, and an agent randomly selects an action with uniform probability if it is feasible from its perspective. The predict node goes through each agent in random order to find an action to simulate forward. If none exists, simulation ends with a score of 0. This means that in the ideal case where all agents share common knowledge of plans, simulation always returns a score of 1.

\paragraph{Back Up}
We take a more customized approach to computing the utility score of each node during the back-up phase. The split node takes the minimum score of the children predict nodes since it cares about the worst-case outcome, similar to the work of \cite{reifsteck2019epistemic}, then multiplies it by the penalty factor of the action that leads to the split node.

The decision node computes the expected utility of the agent's action (including noop) towards contributing to the progression of the task from the perspective of the parent predict node, denoted by $\mathop{\mathbb{E}}_a$ for agent $a$. Given that the subjective (objective) score of an action $act$ is $sc_{act}$ ($oc_{act}$) and the set of best actions for agent $a$ is $Act$, the probability of action $act \in Act$ being taken, denoted by $P_a(act)$, is ${sc_{act}}/{\sum_{act' \in Act}sc_{act'}}$. The utility score of the decision node is then $\sum_{act \in Act} P_a(act) \cdot oc_{act}$. 
Additionally, we set the objective score of the noop action to 0, since it does not contribute to the progression of the task.

The score of the predict node is computed as:
$$ \left(1 - \prod_{a \in Ag}P_a(noop)\right)  \left( \sum_{a\in Ag} \frac{\mathop{\mathbb{E}}_a}{\sum_{i\in Ag} 1-P_i(noop)} \right),$$ which is the probability of at least some agent will act, multiplied by the expected utility of action taken by the first agent who gets to act, since execution is asynchronous. Given that some agent will act, we assume that the probability of agent $a$ acting first is proportional to its probability of taking a non-noop action, i.e. $1 - P_a(noop)$. Therefore, the expected utility is the sum of the normalized probability of each agent $a$ acting first $\frac{1-P_a(noop)}{\sum_{i\in Ag} 1-P_i(noop)}$ multiplied by the expected utility of agent $a$ taking a non-noop action $\frac{\mathop{\mathbb{E}}_a}{1-P_a(noop)}$. This means that if every agent prefers a noop action, then the predict node has a score of 0, i.e. execution is stuck. Note that in reality, agents may decide to act if nobody else does instead of waiting forever. We do not take into account such interactive behavior, but assume this is a reasonable way to approximate the utility of the predict node. 

\paragraph{Implicit Belief Revision}

We consider explanations to be an \textit{explicit} way of revising others' beliefs. We consider it an \textit{implicit} belief revision when an unexpected execution action or intent announcement action causes a less plausible world of an agent to be promoted to be a most plausible world. We assume agents do not wish to surprise others and penalize an action that causes implicit belief revision with a score of 0. This makes sure that our agent always explains its action before taking it if it is not expected by others. However, during execution, implicit belief revision may still occur, such as when others take an action unexpected by our agent which revises our agent's belief.

\paragraph{Performance Optimization} The performance of our algorithm largely depends on the speed of solving constraint satisfaction problems (CSPs) from the knowledge bases. To optimize the performance, we implement incremental checking and caching for CSPs, since the CSPs are largely similar throughout the process. At the decision node, we lazily expand the actions in the order of execution actions, noop action, then communication actions. For example, communication actions do not need to be expanded if higher-priority actions have a better score than the maximum possible score for a communication action due to penalty.

\begin{figure}
    \centering
    \includegraphics[scale=0.50]{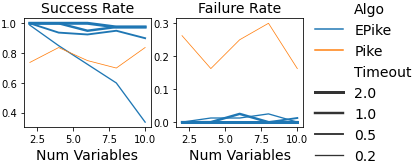}
    \caption{Success rate and failure rate of EPike and Pike on randomly generated test cases}
    \label{fig:exp_1}
\end{figure}

\section{Experiment Results}

We describe our experiment results on the success rate and the scalability of our algorithm {EPike} compared to Pike \cite{levine2018watching}. We implemented our own Pike as a naive version of EPike that assumes what it believes is believed by all, that is, it may falsely assume common knowledge when there is not. Note that the original Pike supports some additional functionalities not accounted for by this paper, such as scheduling. We use z3 as our CSP solver \cite{moura2008z3}. We use an exploration parameter of 4 for \textsc{SearchAction}, and $\sqrt{2}$ for the rest. We use a horizon of $k=3$ for \textsc{SearchAction}. We limit our focus to a 2-agent team. The experiments are run by instantiating two EPike or Pike agents that execute together in a task. We measure the runtime in seconds for one agent being the ego agent, who we assume gets to be the first agent to act if it decides to after each action is taken.

\paragraph{Success Rate}

Since MCTS is an anytime algorithm, we evaluate the success rate and failure rate of EPike and Pike, under different timeout in seconds for MCTS (or if it reaches 1000 iterations of simulations, whichever comes first). The experiments are run for the domains of (1) \textbf{Breakfast}, which includes variations of our motivating example, (2) \textbf{Word Puzzles}, (3) \textbf{Search-and-Rescue} (\textbf{SAR}), (4) Randomly generated sequential tasks. 
We run each hand-crafted test case for 20 times for both Pike and EPike with no timeout, with results shown in Figure \ref{fig:exp_3}. We generate random test cases that vary in the size of the task (number of variables $V$) and the number of constraints that agents differ on, ranging from [0, 3], for 10 cases per condition, and report the result after running each case for 2 times for both Pike and EPike under different timeout, as shown in Figure \ref{fig:exp_1}. Note that it is possible for execution to neither succeed nor fail, in which case execution hangs as no agents plan to act. This could be because (1) MCTS algorithm is stopped by the timeout before it finds a feasible next step, (2) EPike believes execution is bound to fail no matter its action, such as when the other agent would not trust its explanation, or (3) EPike falsely believes that the other agent will act. In practice, we can adopt mitigations such as allowing EPike to take the next best action after having waited for a long time.

From Figure \ref{fig:exp_1}, we see that as timeout increases, EPike's success rate increases, especially for larger-sized tasks, and is higher than Pike's success rate given enough time. Meanwhile, its failure rate is consistently low and always lower than Pike.  This shows that EPike is conservative, and when it does not succeed, it is mainly because it has not found a good action to take within the timeout, but it does not take an incorrect action rashly as Pike tends to do. This is consistent with the result of the hand-crafted test cases in Figure \ref{fig:exp_3}.

\begin{figure}
    \centering
    \includegraphics[scale=0.598]{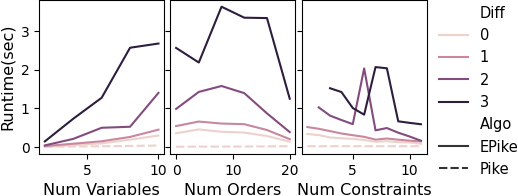}
    \caption{Runtime of EPike and Pike on random tests}
    \label{fig:exp_2}
\end{figure}

\paragraph{Scalability}

To see how EPike scales, we run the MCTS algorithm for a fixed number of 500 iterations to see how long it takes to reach a certain level of certainty under different model parameters, such as the size of the initial plausibility model (embodied by the number of constraints that agents differ on, \textit{Diff} shown by hue of the plot), the size of the task (the number of variables $V$, \textit{Num Variables}), concurrency level of the actions (the number of ordering constraints $O$, \textit{Num Orders}), and the number of constraints $C$ in the task (\textit{Num Constraints}). We measure the average runtime for each callback for Pike and EPike. As shown in Figure \ref{fig:exp_2}, runtime for EPike is heavily affected by how much agents' beliefs differ, and it also increases as the task size increases. Pike takes less time than EPike, as expected, but when under common knowledge, EPike's runtime is closer to Pike and can also finish relatively quickly.

\section{Related Work}

\paragraph{Human-Robot Teaming}
Our work is related to human-robot teaming as it considers the collaborative process of humans and robots working together to achieve tasks. Work in this field focuses primarily on recognizing and adapting to humans' intent, and in some cases, communicating about intent, which we inherit in our work. Pike inspired us to take a constraint-based approach for concurrent intent recognition and adaption, in which a library of precompiled plans are encoded in a knowledge base \cite{levine2018watching}. Pike is later extended to a probabilistic setting, called \textit{Riker} \cite{levine2019risk}, where the robot can further ask the human about their intent \cite{broida2021}. Other than constraint-based approaches, work has been proposed using techniques from classical planning and MDP, such as PReTCIL \cite{Freedman2017IntegrationOP} that uses probabilistic plan recognition and NOPA \cite{puig2023nopa} that leverages inverse planning for goal recognition. 
In \cite{unhelkar2020semi, unhelkar2020decision}, a human behavior model is learned through semi-supervised learning and incorporated into the robot's POMDP model that supports bi-directional communication on intent. However, most work assumes common knowledge of the task, as opposed to implementing an explicit Theory of Mind.

% Human-robot teaming considers the collaborative process of humans and robots working together to achieve tasks. Existing approaches exhibit limited reasoning of an explicit Theory of Mind, and focus on recognizing and, in some cases, communicating about intent. Pike, the predecessor of our work, takes a constraint-based approach that allows concurrent intent recognition and adaption in a unified framework \cite{levine2018watching}. Pike is later extended to a probabilistic setting, called \textit{Riker}, where the robot maintains a model of the human's policy as a dynamic influence diagram \cite{levine2019risk}, and the robot can further ask questions to the human about their intent \cite{broida2021}. Using techniques from classical planning, the PReTCIL framework \cite{Freedman2017IntegrationOP} integrates probabilistic plan recognition \cite{Ramrez2010ProbabilisticPR} and classical planning to provide online robot assistance. 
% Within MDP framework, NOPA \cite{puig2023nopa} uses neural networks combined with inverse planning for efficient goal recognition and an MCTS-based planner for planning the robot's assistance to humans in daily household tasks. 
% In \cite{unhelkar2020semi, unhelkar2020decision}, human's behavior is modeled as an Agent Markov Model (AMM) obtained through semi-supervised learning, which is then incorporated into the robot's POMDP model for decision-making. Their approach also  allows bi-directional communication on intent. 

\paragraph{Epistemic Planning} In the field of epistemic planning, there are two main categories of approaches -- the semantic approach based on Dynamic Epistemic Logic (DEL) \cite{bolander2011epistemic, Le2018EFPAP, Fabiano2020EFP2A} and the symbolic approach \cite{Muise2015PlanningOM}.
Our work leverages the DEL approach and carries over their insight on how to model announcement and question-asking actions.
% Epistemic planning combines classical planning with epistemic logic and solves the problem of automated planning from one agent's perspective to achieve some epistemic goal in a multi-agent system. 
In particular, we take an implicit coordination approach, following from the work of Engesser et al., where the agent takes into account the spontaneous cooperation of other agents in achieving the goal, which requires recursive perspective-taking in order to predict their actions \cite{engesser2017cooperative}. In \cite{bolander2018better}, the authors further discussed the impact of eager and lazy agents in the framework, and in \cite{reifsteck2019epistemic}, an MCTS algorithm is developed that shares similar insights as our work. Compared to the work by Engesser et al., we differ in that our framework based on conditional doxastic logic allows the modeling of false beliefs and the revision of false beliefs, and our explanations refer directly to the plan space instead of states as a result of extending the logic to knowledge bases.

\paragraph{XAIP}

% Work in Explainable AI in Planning (XAIP) aims to increase AI transparency and ensures the robot's plans are explainable to humans.
Our work is related to Explainable AI in Planning (XAIP), especially to the work on plan explanations taking into account the differences in agents' mental models.
In \cite{Chakraborti2017PlanEA}, model reconciliation is proposed that allows robots to explain the model differences upon misalignment between the human's mental model of the robot and the robot's actual model. In \cite{vasileiou2022logic}, a logic-based approach to model reconciliation is proposed, where the planning problem is encoded as a SAT problem using SatPlan, and the model differences are computed with respect to the human and the robot's knowledge bases. Since these approaches consider the entire PDDL planning model, plan explanations go beyond explaining about the differences in the initial states but can also be about agents' discrepancies in goal states and action models.
In \cite{tathagata2019balance}, model reconciliation is balanced with explicable planning, which allows robots to find (potentially sub-optimal) plans that are expected by the human based on the human's understanding of the robot \cite{zhang2017plan}, and in \cite{sreedharan2020expectation}, the two are unified in an expectation-aware planning framework with additional explanatory actions. This inspired us in thinking about how the robot can balance its adaptation and communication with the human.  However, most of their work considers humans as observers without much human-robot cooperation. In \cite{zahedi2022mental}, the authors pointed out the importance of a richer mental modeling framework that allows human-robot collaboration, which we provide a viable way of filling the gap. 

Another line of work from Shvo et al. \cite{shvo2020towards} provides explanations by considering agents' Theory of Mind represented using epistemic logic. In particular, to resolve the human's misconceptions about plans, a symbolic epistemic planner RP-MEP \cite{Muise2015PlanningOM} is used for the robot to either take actions to align the true state to the human's belief or explain the true state to the human \cite{shvo2022resolving}. However, their explanations are also on states rather than plans.

\section{Conclusion}
In this work, we combine insights from epistemic logic and knowledge-base encoding of plans to allow agents to understand  discrepancies in their beliefs of feasible plans. We develop an online execution algorithm Epistemic Pike for the agent to dynamically plan its actions to adapt to others and communicate to resolve any discrepancy. We show that our agent is effective in working in teams where a shared mental model of plans cannot be guaranteed. A natural next step is to consider cases where actions are partially observable.

% \fontsize{9.6pt}{10.6pt} \selectfont

\section{Acknowledgements}

This material is based upon work supported by the Defense Advanced Research Projects Agency (DARPA) under Contract No. HR001120C0035. Any opinions, findings and conclusions or recommendations expressed in this material are those of the author(s) and do not necessarily reflect the views of the Defense Advanced Research Projects Agency (DARPA).

\bibliography{aaai23}

\end{document}